# On Node Features for Graph Neural Networks

**Chi Thang Duong**
EPFL
*thang.duong@epfl.ch*

**Thanh Dat Hoang**
HUST
*thanhdath97@gmail.com*

**Ha The Hien Dang**
Google
*dang.hien@google.com*

**Quoc Viet Hung Nguyen**
Griffith University
*henry.nguyen@griffith.edu.au*

**Karl Aberer**
EPFL
*karl.aberer@epfl.ch*

## Abstract

Graph neural network (GNN) is a deep model for graph representation learning. One advantage of graph neural network is its ability to incorporate node features into the learning process. However, this prevents graph neural network from being applied into featureless graphs. In this paper, we first analyze the effects of node features on the performance of graph neural network. We show that GNNs work well if there is a strong correlation between node features and node labels. Based on these results, we propose new feature initialization methods that allows to apply graph neural network to non-attributed graphs. Our experimental results show that the artificial features are highly competitive with real features.

## 1 Introduction

Graphs are a natural representation of relationships between entities in a system such as social networks or information networks. Representation learning on graphs is increasingly popular as it is able to achieve state-of-the-art results on different learning tasks such as node classification, graph classification or link prediction. Its goal is to construct a low-dimensional embedding space that can capture the graph structure.

There are two main approaches in graph representation learning which are shallow models and deep models. In shallow model such as DeepWalk[11] or node2vec[4], the embeddings are fed directly to a loss function without considering the graph structure or node features. On the other hand, deep models, which are called Graph Neural Networks (GNN), allow to take into account both the graph structure and the node features. However, one advantage of GNN, which is its ability to integrate node features and graph structure into the learning process, also hinders its applicability to several real-world graphs where node features are not available. Graphs in the wild may not contain information about the nodes due to privacy concerns or difficulty in collecting node features.

Existing GNN techniques use different methods to initialize node features as a way to handling featureless graphs. However, these techniques are usually adhoc such as random, constant degree-based initialization. To the best of our knowledge, there is no study on the effects of node features on the performance of GNNs and how GNNs can be used for non-attributed graphs. To this end, in this paper, we conduct a study of different existing feature initialization methods. In addition, by comparing performance using artificial features against real features on attributed graphs, we aim to shed light on the importance of node features on GNN techniques.

## 2 Feature Initialization Methods

In the following, we discuss several initialization methods that have been used in the literature. These techniques can be classified into 2 categories: centrality-based and learning-based. Traditionally, centrality-based techniques are mostly used in the literature as they are easy and fast to compute. In this paper, we propose to initialize node features using learning-based approaches which construct node features based on graph structure using unsupervised node embedding methods.

### 2.1 Centrality-based approaches

These approaches compute a node feature based on its role in the graph. It is worth noting that the features constructed by these methods are usually a scalar number. *Degree* [5]: this is the most popular intitialization method in which a node's feature is its degree. *PageRank* [12]: the pagerank of a node in the graph is used as the node feature. *Egonet* [12, 6]: the feature of a node is the number of edges within the egonet of the node and the number of links adjacent to the egonet. *Number of triangles* [6]: a node's feature is the number of triangles the node participates in. *k-core number* [12]: A k-core is defined to be the largest subgraph such that all the nodes in the subgraph have degree of k or more. Then, the core number of a node is the largest value k among all k-cores which contain that node. *Local coloring number* [12]: is the index of the color in the smallest coloring of the graph. A coloring of a graph is an assignment of colors to nodes such that no adjacent nodes have the same color. *Largest clique number* [12]: the feature of a node is defined as the largest k of a k-clique that contains that node.

### 2.2 Learning-based approaches

Unlike centrality-based approaches in which a node feature is constructed based on its local neighborhood. In learning-based approaches, a node feature is considered as the node embedding obtained from an unsupervised learning process which considers the whole graph structure. In our paper, we consider three representative unsupervised embedding methods which are DeepWalk [11], HOPE [10]. DeepWalk is a shallow embedding approach in which two nodes are considered to be close if they cooccur on a random walk starting from one node. HOPE can be considered as a graph factorization method in which variants of the adjacency matrix of a graph are factorized to obtain the node embeddings.

## 3 Experiments

### 3.1 Experiment settings

**Datasets.** We consider 3 types of datasets depending on the learning task and the availability of node features. For node embeddings, we consider three dataasets without node features which are BlogCatalog[14], Wiki[4] and PPI[4]. We also consider 5 datasets with node features: Cora[13], Citeseer[13], Pubmed[8], Reddit[5] and NELL[17]. For graph embeddings, we consider 5 datasets MUTAG[2], ENZYMES[1], DD[3], COLLAB[16], IMDB[16] in which 3 datasets have node features. The statistics of the datasets are shown in Table 6.

### 3.2 Importance of node features

In this experiment, we shuffle the node features among nodes while keeping the node labels intact. By comparing the difference in performance between results with shuffling and without shuffling, we can measure the importance of node features in comparison with graph structure. If the node features are more informative, the difference would be large. In addition, we expect that GNN techniques that rely on node features more than graph structure will suffer more.

The experimental results are shown in Table 1. We observe that as expected, no shuffling is better than with shuffling. However, the margin between with and without shuffling is different across datasets and techniques. GraphSAGE suffers the most when shuffling is on as it relies heavily on node features to compute the node embeddings while paying less attention to graph structure. Other techniques are less susceptible as graph structure are equally important as node features. For instance, DGI relies on subgraph patches which can capture the graph structure. Among datasets, we observe two different categories: the Planetoid datasets (Cora, Citeseer, Pubmed) and Reddit, NELL. For Planetoid datasets, we observe a large margin of difference between shuffling and without shuffling while for Reddit and NELL datasets, the margin is smaller. For instance, the difference is at most 0.2 for Reddit and NELL but it is at most 0.4 for Planetoid datasets. The reason is that the node labels of Planetoid datasets are created based on the node features. When the labels are not correlated with the node features due to shuffling, the performance of GNN methods on these datasets decreases.

Table 1: Effects of feature shuffling on GNN performance

|  | Shuffle | Cora | Citeseer | Pubmed | Reddit | NELL |
|---|---|---|---|---|---|---|
| GraphSAGE | No | 0.796±0.007 | 0.657±0.006 | 0.766±0.007 | 0.651±0.001 | 0.637±0.015 |
|  | Yes | 0.568±0.004 | 0.387±0.015 | 0.403±0.010 | 0.091±0.002 | 0.085±0.020 |
| DGI | No | 0.812±0.009 | 0.671±0.011 | 0.776±0.013 | 0.944±0.000 | 0.691±0.005 |
|  | Yes | 0.389±0.009 | 0.263±0.008 | 0.374±0.022 | 0.786±0.012 | 0.818±0.006 |
| SGC | No | 0.793±0.002 | 0.674±0.001 | 0.793±0.000 | 0.947±0.000 | 0.810±0.002 |
|  | Yes | 0.441±0.014 | 0.354±0.009 | 0.344±0.003 | 0.770±0.006 | 0.798±0.008 |

### 3.3 Node classification

In this experiment, we compare different initialization strategies on the transductive setting. The experiment was conducted on all node embedding datasets with Simplifing Graph Convolutional Networks (SGC)[15] as the GNN embedding method. The experimental results are shown in Table 2 and 3.

We observe that centrality-based methods are outperformed by learning-based methods significantly. For instance, on the Cora dataset, the highest micro-F1 score among centrality methods is 0.203 which is 0.4 lower than the lowest micro-F1 score among learning-based methods. This is expected for the transductive setting as learning-based initialization methods are able to capture the whole graph structure. However, both approaches tend to perform worse than real features. For instance, real features outperform synthetic features on 3 out of 5 datasets which are Cora, Citeseer and Reddit. This confirms that there is a strong correlation between real features and node labels on these datasets. Similar to the above experiment, we observe that the results on the NELL dataset are different from others. Real features are outperformed by learning-based features obtained from DeepWalk. This also comes from the fact that the graph structure are more informative for NELL dataset.

We compare learning-based methods on datasets without features on the transductive setting. The results are shown in Table 3. We observe similar results as the learning-based approaches outperform centrality-based methods significantly. For instance, on the Wiki dataset, learning-based approaches achieve at least 0.6 in term of micro-F1 score while the highest one from centrality-based methods is only 0.176. This is because centrality-based methods focus on capturing local neighborhood around a node, which is not suitable for the transductive node embedding setting.

We also compare against methods where no aggregation using GNN are used. It is worth noting that using GNN on top of learning-based node features is *better* than using learning-based node features alone. For example, GNN + learning-based node features outperforms learning-based node features only on both Wiki and PPI datasets while the difference on the BlogCatalog is small. This is significant as it shows synthetic features are useful when used in combination with GNN methods. In addition, it also shows that it is possible to use GNN methods for featureless graphs by initializing node features using learning-based methods.

### 3.4 Graph classification

Table 4 shows the experimetal results on graph classification on different datasets with SGC as the embedding technique. A noteworthy observation is that real features perform worse than artificial features in 2 out of 3 attributed datasets which are DD and MUTAG. This shows that the structure information are more important than node features on these datasets. To confirm this hypothesis, we also compare with a simple baseline graph embedding technique in which we average the node embeddings obtained from DeepWalk or HOPE as the graph embedding (the last two rows in Table 4). We observe that such simple strategy is already powerful as it is able to outperform all other techniques on ENZYMES and MUTAG dataset. This can be attributed to the simplicity of these two datasets, which have been discussed in the literature[9].

Table 2: Node classification on attributed datasets (micro-F1 scores)

|  | Cora | Citeseer | Pubmed | Reddit | NELL |
|---|---|---|---|---|---|
| degree | 0.156±0.061 | 0.284±0.013 | 0.416±0.022 | 0.144±0.043 | 0.126±0.000 |
| #triangles | 0.142±0.043 | 0.275±0.000 | 0.484±0.000 | N/A | 0.026±0.000 |
| k-core no. | 0.166±0.077 | 0.235±0.021 | 0.448±0.000 | 0.115±0.019 | 0.200±0.000 |
| egonet no. | 0.180±0.049 | 0.290±0.014 | 0.476±0.002 | N/A | 0.126±0.000 |
| pagerank | 0.146±0.032 | 0.230±0.017 | 0.419±0.003 | 0.147±0.000 | 0.126±0.000 |
| coloring no. | 0.203±0.036 | 0.257±0.006 | 0.421±0.019 | 0.135±0.000 | 0.179±0.000 |
| real features | 0.736±0.005 | 0.635±0.009 | 0.707±0.016 | 0.947±0.000 | 0.833±0.000 |
| DeepWalk | 0.728±0.013 | 0.492±0.021 | 0.731±0.026 | 0.931±0.000 | 0.844±0.004 |
| HOPE | 0.620±0.003 | 0.404±0.010 | 0.672±0.010 | 0.922±0.000 | 0.685±0.005 |

Table 3: Node classification on non-attributed datasets

| | | BlogCatalog | | Wiki | | PPI | |
|---|---|---|---|---|---|---|---|
| | | Micro-F1 | Macro-F1 | Micro-F1 | Macro-F1 | Micro-F1 | Macro-F1 |
| SGC | degree | 0.167±0.005 | 0.027±0.003 | **0.176±0.018** | **0.030±0.002** | 0.077±0.002 | 0.026±0.001 |
| | #triangles | 0.164±0.001 | 0.026±0.001 | 0.139±0.012 | 0.027±0.003 | 0.079±0.004 | 0.029±0.004 |
| | k-core no. | 0.161±0.004 | 0.027±0.002 | 0.127±0.010 | 0.023±0.002 | **0.090±0.004** | **0.032±0.002** |
| | egonet no. | 0.165±0.009 | 0.025±0.002 | 0.155±0.038 | 0.029±0.005 | 0.078±0.003 | 0.024±0.002 |
| | pagerank | **0.168±0.005** | 0.026±0.002 | 0.161±0.019 | 0.019±0.002 | 0.077±0.002 | 0.024±0.002 |
| | coloring no. | 0.163±0.005 | **0.028±0.002** | 0.141±0.014 | 0.027±0.002 | **0.090±0.002** | 0.031±0.002 |
| | DeepWalk | 0.312±0.008 | 0.106±0.002 | **0.685±0.019** | **0.572±0.028** | 0.256±0.004 | **0.191±0.006** |
| | HOPE | **0.320±0.005** | **0.118±0.003** | 0.636±0.016 | 0.509±0.014 | 0.223±0.002 | 0.143±0.002 |
| None | DeepWalk | **0.385±0.005** | **0.223±0.007** | 0.623±0.011 | 0.522±0.036 | 0.222±0.010 | 0.188±0.003 |
| | HOPE | 0.321±0.003 | 0.144±0.007 | 0.597±0.012 | 0.491±0.021 | 0.191±0.004 | 0.145±0.006 |

Regarding artificial features, centrality-based methods especially degree-based initialization are better than learning-based ones. For instance, the difference between degree-based features and the best of learning-based ones are at least 0.03 on IMDB dataset. This can be explained by how the datasets are constructed. For both Collab and IMDB, the graphs are egonets around nodes from 3 larger graphs. As the labels of the graphs come from the labels of the nodes, the graph structure are more useful for classification than the node features.

## 4 Conclusion

We have performed an experimental study to compare different feature initialization methods for non-attributed graphs. Our experiments also showed that GNN only work well if there is a strong correlation between node features and node labels. In addition, we showed that traditional centrality-based methods are outperformed by learning-based methods such as DeepWalk. Moreover, in many cases, we observe that artificial features even outperform real features and the combination of artificial features and GNN is also helpful in many settings.

Table 4: Graph classification (micro-F1 scores)

|  |  | ENZYMES | DD | MUTAG | Collab | IMDB |
|---|---|---|---|---|---|---|
| SGC | degree | **0.335±0.051** | 0.720±0.035 | **0.860±0.058** | **0.778±0.005** | **0.732±0.033** |
|  | #triangles | 0.230±0.038 | 0.569±0.000 | 0.719±0.062 | 0.758±0.019 | 0.712±0.008 |
|  | k-core no. | 0.276±0.043 | **0.763±0.000** | 0.830±0.051 | 0.743±0.000 | 0.700±0.000 |
|  | egonet no. | 0.295±0.032 | 0.737±0.000 | 0.860±0.073 | 0.773±0.000 | 0.700±0.010 |
|  | pagerank | 0.307±0.045 | 0.748±0.007 | 0.850±0.045 | 0.774±0.036 | 0.688±0.002 |
|  | coloring no. | 0.242±0.000 | 0.737±0.000 | 0.859±0.047 | 0.770±0.013 | 0.677±0.055 |
|  | real features | 0.393±0.040 | 0.698±0.009 | 0.810±0.086 | N/A | N/A |
|  | DeepWalk | **0.230±0.019** | **0.731±0.014** | 0.810±0.073 | 0.597±0.000 | **0.700±0.030** |
|  | HOPE | 0.181±0.014 | 0.711±0.000 | **0.812±0.031** | **0.640±0.000** | 0.617±0.013 |
| None | DeepWalk | **0.279±0.021** | **0.715±0.004** | **0.892±0.054** | **0.596±0.041** | **0.573±0.017** |
|  | HOPE | 0.188±0.029 | 0.611±0.020 | 0.635±0.014 | 0.524±0.000 | 0.480±0.020 |

Table 5: Node embedding datasets

| Dataset | $|V|$ | $|E|$ | $k$ | #Classes |
|---|---|---|---|---|
| BlogCatalog | 10'312 | 333'983 | N/A | 39 |
| Wiki | 4'777 | 184'812 | N/A | 40 |
| PPI | 3'890 | 76'584 | N/A | 50 |
| Cora | 2'708 | 5'429 | 1'433 | 7 |
| Citeseer | 3'327 | 4'732 | 3'703 | 6 |
| Pubmed | 19'717 | 44'338 | 500 | 3 |
| Reddit | 231'443 | 11'606'919 | 602 | 41 |
| NELL | 65'755 | 266'144 | 5'414 | 210 |

Table 6: Graph embedding datasets

| Dataset | Avg. $|V|$ | Avg. $|E|$ | #Graphs | $k$ | #Classes |
|---|---|---|---|---|---|
| MUTAG | 17.93 | 19.79 | 188 | 1 | 2 |
| ENZYMES | 32.63 | 62.14 | 600 | 18 | 6 |
| DD | 284.32 | 715.66 | 1'178 | 1 | 2 |
| COLLAB | 74.49 | 2'457.78 | 5'000 | N/A | 3 |
| IMDB | 13 | 65.94 | 1'000 | N/A | 3 |

## A  Additional dataset details

BlogCatalog[14] is a social network of bloggers in which the node labels capture the topics of interest of the bloggers. Wiki[4] is the word adjacency graph of the first million bytes of the Wikipedia dump from text8[7]. The labels of the words are their part-of-speech tags. PPI[4] is a subgraph extracted from the PPI network for human and the labels represent the biological states.

For node embeddings, Cora[13], Pubmed[8] and Citeseer[13] are three bibliographic information networks in which the nodes are papers and the edges are referencing relationships. Node labels represent the domains of the papers while the features are the presence/absence of some keywords. The Reddit[5] dataset is a network of reddit posts where there is an edge between two posts if they have the same author. The labels of the posts are the subreddits they belong to. A post's feature is constructed from the word embeddings of its title, comments and the number of comments it received. NELL[17] is an entity classification dataset in which the nodes are the entities of a knowledge graph. The labels are entity types and the features are bag-of-words representing the entities. The statistics of the node embedding datasets are shown in Table 5.

MUTAG[2], ENZYMES[1] and DD[3] are protein datasets in which each graph is the molecule structure of a protein. COLLAB[16] is a scientific collaboration dataset obtained by combining from 3 collaboration datasets belonging to 3 different fields. A graph in COLLAB is an egonet of a researcher and its label is the field of the researcher. Similarly, IMDB[16] is a movie-collaboration dataset where each graph is an egonet of an actress/actor. The label of an egonet is the genre of the movie the actor/actress performed in.